\documentclass[10pt, a4paper]{article}
\usepackage{svg}
\usepackage{amsmath}
\usepackage{amsfonts}

\usepackage[final]{lrec2026} 

\usepackage{arydshln}

\newcommand{\R}{\mathbb{R}}

\title{Creating a Hybrid Rule and Neural Network Based Semantic Tagger using Silver Standard Data: the PyMUSAS framework for Multilingual Semantic Annotation}

\name{Andrew Moore\thanks{$^{*}$Corresponding authors. $^{\dagger}$Johanna Vuorinen and the referenced Laura Löfberg refer to the same individual.}$^{*1}$, Paul Rayson$^{*1,8}$, Dawn Archer$^2$, Tim Czerniak$^3$, Dawn Knight$^4$, \\ {\bf \large Daisy Lal$^1$, Gearóid Ó Donnchadha$^5$, Mícheál Ó Meachair$^6$, Scott Piao$^1$,} \\ {\bf \large Elaine Uí Dhonnchadha$^3$, Johanna Vuorinen$^{\dagger5}$, Yan Yabo$^7$, Xiaobin Yang$^7$}}

\address{$^1$UCREL, Lancaster University, UK; $^2$Manchester Metropolitan University, UK;\\$^3$Centre for Language and Communication Studies, Trinity College, Dublin, Ireland;\\$^4$School of English, Communication and Philosophy, Cardiff University, Wales;\\$^5$Independent Researcher;\\$^6$Fiontar \& Scoil na Gaeilge, Dublin City University, Ireland;\\$^7$Hubei University, China; $^8$VinUniversity, Vietnam\\{a.p.moore, p.rayson}@lancaster.ac.uk}

\abstract{
Word Sense Disambiguation (WSD) has been widely evaluated using the semantic frameworks of WordNet \citep{maru-etal-2022-nibbling}, BabelNet \citep{Pasini_Raganato_Navigli_2021}, and the Oxford Dictionary of English \citep{gadetsky-etal-2018-conditional, Chang2018xSenseLS}. However, for the UCREL Semantic Analysis System (USAS) framework, no open extensive evaluation has been performed beyond lexical coverage or single language evaluation. In this work, we perform the largest semantic tagging evaluation of the rule based system that uses the lexical resources in the USAS framework covering five different languages using four existing datasets and one novel Chinese dataset. We create a new silver labelled English dataset, to overcome the lack of manually tagged training data, that we train and evaluate various mono and multilingual neural models in both mono and cross-lingual evaluation setups with comparisons to their rule based counterparts, and show how a rule based system can be enhanced with a neural network model. The resulting neural network models, including the data they were trained on, the Chinese evaluation dataset, and all of the code have been released as open resources.  
 \\ \newline  }

\begin{document}

\maketitleabstract

\section{Introduction and Related Work}

Word Sense Disambiguation (WSD) is the task of assigning a word with a pre-defined sense inventory according to a given context. WSD as a field has progressed significantly from early uses of feature-based Support Vector Machines (SVMs) per word-type\footnote{Word-type here refers to the word and its Part Of Speech.} \citep{zhong-ng-2010-makes} to more recent fine-tuning of Pre-trained Language Models (PLMs) \citep{barba-etal-2021-esc}. Current Large Language Models (LLMs) \citep{basile2025exploring}, with prompting of LLMs \citep{Meconi2025DoLL, basile2025exploring} achieve State-Of-The-Art (SOTA) performance on various evaluation setups. In addition to the development of models, WSD has grown with respect to datasets for both training (e.g. SemCor and the WordNet Gloss Corpus) \citep{miller-etal-1994-using, langone-etal-2004-annotating} and evaluating \citep{raganato-etal-2017-word, maru-etal-2022-nibbling} on English as well as training \citep{conia-etal-2024-mosaico} and evaluating \citep{Pasini_Raganato_Navigli_2021} on a variety of other languages. However, all of this has been created within the WordNet \citep{10.1145/219717.219748} or BabelNet \citep{ijcai2021p620} semantic frameworks.

In this work, we focus on evaluating and extending semantic tagging tools and datasets within the USAS semantic framework \citep{rayson2004ucrel} which, compared to WordNet and BabelNet, provides a more coarse-grained sense inventory, as shown in Table \ref{table:sense_inventory_sizes}\footnote{Statistics for WordNet taken from: \url{https://wordnet.princeton.edu/documentation/wnstats7wn}}\footnote{Statistics for BabelNet taken from: \url{https://babelnet.org/statistics}}. The framework is language independent, similar to other semantic frameworks such as BabelNet and Open Multilingual WordNet \citep{bond-foster-2013-linking}.

Most of the previous work within the USAS framework has focused on developing rule-based semantic taggers, which at their core rely on language-specific semantic lexicons, including the original C version of the English semantic tagger \citep{rayson2004ucrel}\footnote{C is the programming language that the original English rule based tagger was written in.}. More recently, the system has been re-developed and expanded to 10 languages through a Python based framework called PyMUSAS\footnote{\url{https://github.com/UCREL/pymusas}} that utilises language specific lexicons which have been developed manually and semi-automatically through previous works and a large community effort \citep{piao-etal-2015-development, piao-etal-2016-lexical}.

In this research, we demonstrate the power of combining neural networks and the USAS rule-based model to create a novel hybrid model for semantic tagging. We extend the work of \citet{ezeani-etal-2019-leveraging} for the Welsh language by fine-tuning larger pre-trained models, specifically Pre-trained Language Models (PLMs). The PLMs were fine-tuned on over 5 million English-only silver labelled tokens generated by the English rule-based tagger, thus alleviating the need for a manually annotated training dataset. As the PLMs are both English and multilingual based, in addition to the Welsh neural tagger, we produce the first neural network models trained specifically for USAS tagging for English, Irish, Finnish and Chinese.

We collate existing manually annotated datasets across the four different languages, with an additional newly labelled Chinese dataset, to generate a new benchmark covering the five languages, allowing us for the first time to compare rule-based, neural network, and hybrid models. In addition, this is the first time models have been contextually evaluated for USAS tagging beyond a single language with open datasets \citep{piao-etal-2015-development, ezeani-etal-2019-leveraging, czerniak-ui-dhonnchadha-2024-towards}. All data within this paper have been made publicly available or are accessible upon request\footnote{The Irish data that is used for evaluation in this paper is available upon request, please email your request to: \href{mailto:UIDHONNE@tcd.ie}{Dr. Elaine Uí Dhonnchadha}}.

\begin{table}[!ht]
    \centering
    \begin{tabular}{|c|c|}
        \hline
        Sense Inventory & No. Senses \\
        \hline
        USAS & 232 \\
        WordNet 3.0 & 117,659 \\
        BabelNet v4.0 & 15,780,364 \\
        BabelNet v5.3 & 22,892,310 \\
        \hline
    \end{tabular}
    \caption{The number of unique senses in each sense inventory (No. Senses), also known as synsets, for each sense inventory. WordNet 3.0 is displayed here as it is the sense inventory used in the most popular benchmarking datasets \citep{raganato-etal-2017-word, maru-etal-2022-nibbling} and BabelNet v4.0 and v5.3 are shown as they are used in a popular multilingual benchmarking dataset \citep{Pasini_Raganato_Navigli_2021} and is the latest version respectively.}
    \label{table:sense_inventory_sizes}
\end{table}

The main contributions of this work include:
\begin{enumerate}
    \item The first neural network based English and multilingual semantic taggers that have been trained specifically for the USAS tagset on English silver labelled data.
    \item A demonstration of how a neural model can enhance an existing rule-based system, through the creation of a hybrid rule/neural based model.
    \item An evaluation of the contextual correctness of tags in the existing rule-based systems for four languages for the first time.
    \item The first comparison of rule, neural, and hybrid systems in five languages.
    \item The release of the first open access manually annotated corpus for USAS semantic tagging in Chinese.
\end{enumerate}

All of the resources mentioned in the paper have been released open access with the publication of this paper, or will be available on request\footnote{Code for training the models; \url{https://github.com/UCREL/experimental-wsd} and the trained models, training dataset, and evaluation datasets; \url{https://huggingface.co/collections/ucrelnlp/usas-neural-taggers-10}}.

\section{USAS Tagset}

The USAS tagset \citep{rayson2004ucrel, lofberg2019developing}, which was originally derived from the Longman Lexicon of Contemporary English \citep{longman1981}, contains 21 major discourse fields that expand to 232 category labels. USAS has a hierarchical tagset that has up to three levels of sub-division. An example of the tagset is shown in Table \ref{table:selection_usas_tags}. These 232 category labels can be enhanced with affixed symbols to indicate additional linguistic meaning, such as rarity (\%@), gender (mf), antonyms (+-), and can be applied to both single words and Multi-Word Expressions (MWE). The semantic tags can be combined with double or triple membership through the use of a slash ("/"). For instance, in Table \ref{table:example_usas_texts} the token "\textbf{Coffee}" is tagged with "\textit{F2/O2}", indicating the token is related to both Drinks (\textit{F2}) and Objects (\textit{O2}), which is correct. Examples of semantically tagged tokens with the full USAS tagset can be seen in table \ref{table:example_usas_texts}, and a full explanation of the USAS tagset can be found in \citet{archer2002introduction}.

In this work, we focus on evaluating on the 232 category labels without any of the affixed symbols nor modelling the MWE. Unlike the rule-based models, the neural models were trained to predict the 232 category labels without the dual or triple tag membership, in order to simplify the training procedure as it reduces the number of category labels\footnote{For dual membership the number of category labels increase by the power of 2 and for triple membership by the power of 3.} and reduces the number of training examples per category.

As the neural model requires a description of each category label, also known in WSD as the gloss, we used the combination of the title and if one exists description of the tag from \citet{archer2002introduction} as the gloss.

\begin{table*}[!ht]
    \centering
    \begin{tabular}{c}
         \hline
         \textbf{Oxygen} \textit{\{O1.3\}} \textbf{,} \textit{\{PUNC\}} \textbf{light} \textit{\{O4.3\}} \textbf{and} \textit{\{Z5\}} \textbf{moisture} \textit{\{O1.2\}} \\
         \textbf{Coffee} \textit{\{F2/O2[i135.2.1\}} \textbf{pot} \textit{\{F2/O2[i135.2.2\}} \\
         \textbf{Erik} \textit{\{Z1mf\}} \textit{\{Z3c\}} \textbf{Adolf} \textit{\{Z1mf\}} \textit{\{Z3c\}} \textbf{von} \textit{\{Z1mf\}} \textit{\{Z3c\}} \textbf{Willebrand} \textit{\{Z99\}} \\
         \hline
    \end{tabular}
    \caption{Example of \textbf{tokens} semantically annotated with USAS tags. Each token is tagged with one or more \textit{USAS tag groups}, indicated by the curly braces (\{\}). The "PUNC" or sometimes seen as "PUNCT" indicates a punctuation token and is not part of the USAS tagset. The first two lines come from a coffee related article and the last comes from the Wikipedia article on Erik Adolf von Willebrand.}
    \label{table:example_usas_texts}
\end{table*}

\begin{table}[!ht]
    \centering
    \begin{tabular}{|c|p{5.7cm}|}
        \hline
        Tag & Short Description \\
        \hline
        F2 & Drinks \\
        O1.2 & Substances and materials generally: Liquid \\
        O1.3 & Substances and materials generally: Gas \\
        O2 & Objects generally \\
        O4.3 & Colour and colour patterns \\
        Z1 & Personal names \\
        Z3 & Other proper names \\
        Z5 & Grammatical bin \\
        Z99 & Unmatched \\
        \hline
    \end{tabular}
    \caption{A selection of USAS tags with a short description.}
    \label{table:selection_usas_tags}
\end{table}

\section{Dataset}

In this section, we detail the silver-labelled training data that we have created for the neural based models and the evaluation data.

\subsection{Training Data}
\label{section:silver_training_data}

Inspired by \citet{conia-etal-2024-mosaico} who created a large silver labelled dataset named MOSAICo that was used for WSD, Semantic Role Labelling, Semantic Parsing, and Relation Extraction, we have also created a silver labelled dataset for WSD using the USAS tagset instead of the BabelNet tagset. 

In comparison to MOSAICo, where they used pre-trained models that have already been trained on a smaller manually annotated dataset (229,517 + 496,776 annotated words from SemCor and WordNet GlossTag corpora \citep{vial-etal-2018-ufsac}), we used the original rule-based English-only C version of the semantic tagger \citep{rayson2004ucrel} for creating the silver standard training data. The C version of the tagger was used as it is more complex and accurate compared to the newer PyMUSAS version, however we use the PyMUSAS version later in the paper to evaluate on and compare to other methods  as the C-version is not open source or widely available\footnote{A version of the tagger was given to us from the original authors for this work and have given us permission to distribute the tags generated from the tagger under the same license as the dataset.}.

We also used the higher quality data from the MOSAICo dataset, denoted as \textit{MOSAICo Core}, to create our silver labelled training dataset. This data contains English Wikipedia documents labelled as either "good" or "featured". \citet{conia-etal-2024-mosaico} demonstrated that by using a higher quality but smaller silver training dataset, they could achieve comparable results to a model trained on a large Wikipedia dataset that has not been filtered by quality.

\begin{table*}[!ht]
    \centering
    \begin{tabular}{|c|c|c|c|c|c|c|}
    \hline
         Split & Documents & Sentences & Tokens & L. Tokens & Labels & Labels per Token \\
    \hline
        Train & 1,083 & 444,880 & 6,651,919 & 5,378,222 & 11,362,935 & 2.11 \\
        Validation & 21 & 20,000 & 258,561 & 204,871 & 453,300 & 2.21 \\
    \hline
    \end{tabular}
    \caption{The number of documents, sentences, tokens, Labelled Tokens (L. Tokens), and labels in the silver labelled training dataset per split. Each labelled token could have more than one label associated it hence the labels per token ratio.}
    \label{table:silver_labelled_training_data_statistics}
\end{table*}

The Wikipedia data was tokenized, sentence-split, and POS-tagged by the CLAWS tagger and then lemmatized and semantically tagged by the C-version of the semantic tagger. The composition of this corpus can be seen in table \ref{table:silver_labelled_training_data_statistics}, where we have also divided the data into training and validation splits. The semantic tagger labels each token with one or more USAS tags derived from all possible senses determined by the lexicons. If a tag contains dual or triple membership, we split the tag up into 2 or 3 separate tags respectively. We removed all tags marked as punctuation or containing the unmatched USAS ("Z99"), as these tags do not have any semantic meaning.

As this dataset only contains positive USAS tags per token, whereby all positive senses for the target token $i$ is represented by $S_{pi} = {s_1,...,s_c}$ and all USAS tags are represented by $S$ where $S_{pi} \subset S$, to train the neural model we need negative examples per token as well.

We randomly sampled three negative USAS tags per positive tag from three different weighted distributions; 1. the USAS tag distribution from the silver labelled training data split (\textit{Original}), 2. the inverse frequency of the USAS tag distribution (\textit{Inverse}), and 3. the log to the base 2 of the inverse frequency (\textit{Log Inverse}). The reason for the 3 sampling distributions is to keep a balance between; sampling from the original distribution, sampling labels that are under-represented (\textit{Inverse}), and by sampling \textit{Log Inverse} to keep a more even distribution that is more likely to sample from tags that are neither massively over or under representative. This sampling strategy should, to a point, even out the skewed label distribution that is present in USAS labelled data\footnote{The distribution of the USAS tagset within this silver labelled training dataset split can be seen in 
figure \ref{figure:probability_distribution_of_usas_in_silver_data} in appendix A.}, which is common in most datasets that contain large label sets. This inverse frequency negative sampling strategy was inspired by \citet{blevins-zettlemoyer-2020-moving} who weighted the word sense labels by their inverse frequency within the model's loss function so that the model performed better on less frequent word senses for WSD.

These negative samples, which can be represented as $S_{ni} = {s_1, s_2, s_3}$ for target word $i$, are constrained so that no negative USAS tag is a member of the positive USAS tags for that target word $i$ thus $S_{ni} \cap S_{pi} = \{\}$.

This negative sampling strategy is used for both the training and validation split of the silver training data.

\subsection{Evaluation Data}

All the evaluation data has either been manually tagged or has been manually checked. For English and Finnish, we used the dataset from \citet{lofberg2003porting}, a tagged corpus of texts from a Finnish coffee website\footnote{The website URL: \url{http://www.kahvilasi.net/}. This website is no longer available.}, the English corpus is a machine translated version of the Finnish that was post edited by a native Finnish speaker.

For the Welsh language, we used the entire dataset released in \citet{ezeani-etal-2019-leveraging} for evaluation. The dataset consists of 8 extracts from 4 diverse data sources; Kynulliad3\footnote{\url{https://kevindonnelly.org.uk/kynulliad3/}} (Welsh Assembly proceedings), Meddalwedd\footnote{\url{https://techiaith.cymru/corpws/Moses/Meddalwedd/}} (translations of software instructions), Kwici\footnote{\url{https://kevindonnelly.org.uk/kwici/}} (Welsh Wikipedia articles), and LERBIML\footnote{\url{https://www.lancaster.ac.uk/fass/projects/biml/bimls3corpus.htm}} (multi-domain spoken corpora). In the original work with manually corrected POS tags, 90\% of this dataset was used for training a joint POS and semantic tagger, while 10\% was used for evaluation. 

For the Irish language, we used the data released by \citet{czerniak-ui-dhonnchadha-2024-towards} which contains 3 texts;\footnote{This dataset has been extended since \citet{czerniak-ui-dhonnchadha-2024-towards} to 10 texts, but we decided to use the original 3 so that readers could compare it to the results of the original work that used a different evaluation metric. In the future we will use the extended dataset for evaluation.} a paragraph from an online news article, and the first 226 and 301 tokens of two Wikipedia articles about the TV show \textit{The Wire} and the author George Orwell respectively. The Irish data also has corrected lemma and POS tags.

The Chinese dataset that we have created is a manually tagged text from the ``News Report'' genre of the ToRCH2019 corpus \citeplanguageresource{torch2019}, specifically about the 2019 military world games in Wuhan, China. This dataset was manually annotated following a three-stage procedure; 1. independent tagging, 2. independent reviews of their own tagging, and 3. reaching consensus between two trained researchers. 

Similar pre-processing to the training data was performed on these evaluation datasets; we removed all tag tokens marked as punctuation or containing the unmatched USAS ("Z99"), as these tags do not have any semantic meaning. In addition, any labelled tokens that could not be matched to the USAS tagset were removed. The statistics for these datasets are shown in Table \ref{table:evaluation_data_statistics}\footnote{The Chinese and Irish dataset for 301 and 6 tokens contain more than one USAS label assigned to them, therefore we have used the first USAS label as the only label for these tokens assuming that label is the most likely.}. Unlike the training dataset,  all labelled tokens in the checked evaluation datasets contain only one USAS tag each.

\begin{table*}[!ht]
    \centering
    \begin{tabular}{|c|c|c|c|c|c|}
    \hline
    Language & Text Level & Texts & Tokens & L. Tokens & Multi Tag Membership (\%) \\
    \hline
        Chinese & sentence & 46 & 2,312 & 1,747 & 1 (0\%) \\
        English & sentence & 73 & 3,899 & 3,468 & 212 (6.1\%) \\
        Finnish & sentence & 72 & 2,439 & 2,068 & 254 (12.3\%) \\
        Welsh & sentence & 611 & 14,876 & 12800 & 1,311 (10.2\%) \\
        Irish & paragraph & 3 & 711 & 618 & 55 (8.9\%) \\
        \hline
    \end{tabular}

    \caption{Evaluation dataset statistics. The number of texts, tokens, Labelled Tokens (L. Tokens). Multi Tag Membership is the number of labelled tokens, and the percentage (\%), whereby the USAS tag either has dual, triple, or quadruple membership.}
    \label{table:evaluation_data_statistics}
\end{table*}

\section{Models}

In this section, we outline the rule-based semantic taggers used in this experiment, along with the architecture of the neural based model, and we discuss how the neural model is incorporated into the rule-based model that creates the hybrid model.

\subsection{Rule-Based Models}
\label{section:rule-based-models}

We use the PyMUSAS framework for our rule-based methods which uses one or more lexicons to assign one or more USAS tags to a token. The framework is very flexible, but it works most effectively when additional tools are used in conjunction (e.g. lemmatisers and POS taggers), as these linguistic features are used by the framework to more accurately disambiguate tokens to their respective USAS tag(s) through the lexicons. If a token has more than one USAS tag after other disambiguation methods have been applied, then the first USAS tag is considered the most likely one and the last the least likely one.

The lexicons come in two types, \textbf{single word} and \textbf{MWE}, and both are in essence dictionary lookups. Examples of these lexicons are shown in Tables \ref{table:example_single_word_lexicon} and \ref{table:example_mwe_lexicon} respectively. The \textbf{single word} lexicon matches a word to a list of USAS tags based on its lemma and POS. The \textbf{MWE} lexicon uses a MWE template system (a simplified pattern matching code similar to regular expression) of ``\{token/lemma\}\_\{POS\}'' to match more than one word to a list of semantic tags whereby each word will then be assigned those semantic tags. For example,  the MWE template ``\textit{*\_* Ocean\_NOUN}'' would match to the MWE ``Pacific Ocean'', and both ``Pacific'' and ``Ocean'' would be assigned the same semantic tag of ``Z2''. For some languages, the lexicons have been completely manually created (e.g. English and Finnish), whereas for others they have been automatically created by translation and then partially manually checked to varying degrees (e.g. Welsh, Spanish and Italian).  

\begin{table}[!ht]
    \centering
    \begin{tabular}{|c|c|c|}
        \hline
        Lemma & POS & USAS tags \\
        \hline
        coffee-house & NOUN & H1/F1 \\
        programming & VERB & Y2 P1 \\
        \hline
    \end{tabular}
    \caption{Example single word lexicon where the USAS tags assigned to a lemma and POS are whitespace separated.}
    \label{table:example_single_word_lexicon}
\end{table}

\begin{table}[!ht]
    \centering
    \begin{tabular}{|c|c|}
        \hline
        MWE Template & USAS tags \\
        \hline
        *\_* Ocean\_NOUN	 & Z2 \\
        *\_VERB over\_ADV & T2- M1 M6 \\
        \hline
    \end{tabular}
    \caption{Example MWE lexicon where the USAS tags assigned to a MWE template are whitespace separated.}
    \label{table:example_mwe_lexicon}
\end{table}

The PyMUSAS framework follows a set of heuristics based on those stated on page 4 column 2 of \citet{piao-etal-2003-extracting}\footnote{For a detailed list of these rules see: \url{https://ucrel.github.io/pymusas/api/rankers/lexicon_entry\#contextualrulebasedranker}} which in essence ranks all lexicon matches for a token. For example, \textbf{MWE} matches are ranked higher than \textbf{single word} lexicon matches, and  the search for a match can be expanded through dropping the POS tag requirement or lower casing the token etc.

For each language we evaluate on, we start by using the existing resources and methods with PyMUSAS, i.e. each language has its own rule-based tagger along with the resources, as shown in Table \ref{table:rule_based_tagger_resources_overview}\footnote{All spaCy Transformer and large language models can be found at: \url{https://spacy.io/models}}. The Irish lexicons come from \citet{czerniak-ui-dhonnchadha-2024-towards}\footnote{The Irish tagger from the original paper has been used here. A later improved version will be used in future testing.}, the English, Chinese and Finnish lexicons come from \citet{piao-etal-2016-lexical}, and the Welsh lexicons come from \citet{piao-etal-2018-towards}\footnote{All the lexicons apart from Irish are available from: \url{https://github.com/UCREL/Multilingual-USAS}}.

\begin{table*}[!ht]
    \centering
    \begin{tabular}{c|c|c|c}
        \hline
        Language & Lemmatizer and POS tagger & Single word lexicon & MWE lexicon \\
        \hline
        Irish & \citep{ui-dhonnchadha-van-genabith-2006-part} & 30,190 & 6,825 \\
        English & spaCy Transformer & 54,797 & 19,042 \\
        Chinese & spaCy Transformer & 64,541 & 19,040 \\
        Finnish & spaCy Large & 46,226 & 0 \\
        Welsh & CyTag \citep{neale-etal-2018-leveraging} & 143,292 & 240 \\
        \hline
    \end{tabular}
    \caption{Overview of the resources used for each language's rule-based tagger. When spaCy Transformer or Large is listed this is a language specific spaCy model.}
    \label{table:rule_based_tagger_resources_overview}
\end{table*}

\subsection{Neural Models}
\label{section:neural_models}

We used the WSD Bi-Encoder Model (BEM) from \citet{blevins-zettlemoyer-2020-moving}, where the model is trained to predict the correct sense definition (gloss) for a given ambiguous word given numerous possible sense definitions, where only one of the sense definitions is correct. The architecture of this model is illustrated in Figure \ref{fig:bem_model_architecture}.

In detail, the task is given a text $T = {t_i,...,t_n}$ of $n$ words disambiguate the target word at position $i$ within the text $t_{i}$ to a sense $s_j$ from the set of all possible senses $s_j \in S$ within the given sense inventory $S = {s_1,...,s_m}$, where the sense inventory is the 232 USAS categories. Each sense in the sense inventory will have a corresponding gloss text $G = {g_1,...,g_m}$ describing the sense, each gloss text is represented by a sequence of gloss text tokens $g_n = {gt_1,...,gt_p}$. 

As shown in Figure \ref{fig:bem_model_architecture}, the model encodes the text that contains the target word ($i$)\footnote{If the word is made up of more than one sub-word tokens from the PLM sub-word tokenizer, then the average vector of the sub-word tokens is used to represent the word.} to disambiguate using the context encoder ($C$) which is a Pre-trained Language Model (PLM), creating the vector $u = C(T)$ where $u \in \R^{1\times d}$ and $d$ is the size of the hidden dimension of the PLM.

The text from each gloss is encoded using a PLM from the gloss encoder ($GE$), denoted as $GE_{plm}$. Each gloss text is represented by the vector $j_n \in \R^{1 \times d}$ which is the mean representation of the gloss encoder PLM $j_n = \frac{1}{p}\sum_{i=1}^{p} GE_{plm}(gt_i)[i]$, whereby the gloss encoder and context encoder use the same shared PLM\footnote{In the original BEM model the context and gloss encoders did not share a PLM model, they used independent PLM models. We decided to share the PLM to reduce the number of parameters within the model.}. All of the senses are then represented as the concatenation of all senses' representation denoted by the matrix $J \in \R^{m \times d}$.

The score for each target word is the dot product between the encoded target word and the sense representations $score_i = u \cdot J^T$ where $score_i \in \R^{1 \times m}$, the sense with the highest score is the sense we assign to the target word. The score for a single sense can be represented as $score_{i,n} = u \cdot j_{n}^{T}$.

When training the model, using the silver labelled data described in section \ref{section:silver_training_data}, we use the cross-entropy loss function, as shown in equation \ref{equation:cross_entropy_loss}, using the 3 negative samples per target word $i$. For each sample during training, the model has to predict between 4 senses, 3 are negative and 1 is positive, which is represented by the symbol $v$ in equation \ref{equation:cross_entropy_loss}.

\begin{equation}
\mathcal{L}(t_i) = -log \frac{exp(score_{i,v})}{\sum_{q=1}^{4}exp(score_{i,q})}
\label{equation:cross_entropy_loss}
\end{equation}

\begin{figure*}[!ht]
    \centering
    \includesvg[width=\linewidth]{figures/WSD_BEM_Model.svg}
    \caption{Architecture of the WSD Bi-Encoder Model (BEM) from \citet{blevins-zettlemoyer-2020-moving}.}
    \label{fig:bem_model_architecture}
\end{figure*}

The BEM model was chosen over more accurate and recent models from the WSD literature due to the model's computational efficiency, while the more recent and accurate models \citep{barba-etal-2021-esc, zhang-etal-2022-word} are all cross-encoders. To use cross-encoders at inference time to disambiguate all words in a text would require running the PLM $n \times m$ times, whereby $n$ is the number of words in the given text to disambiguate and $m$ are the number of senses in the sense inventory compared to $m + 1$ for the BEM model\footnote{The BEM model can pre-compute the $m$ sense embeddings in this case disambiguating all words in a given text would only require running the context encoder PLM once.}. The computational efficiency difference between bi-encoders and cross-encoders is highlighted in previous works \citep{reimers-gurevych-2019-sentence,Humeau2019PolyencodersAA}.

\subsection{Hybrid Model}

The major limitation of the rule-based models is lexical coverage, that is, the extent to which words in a text are recognised. This coverage is determined by the lexicon size; in general, the larger the lexicon, the greater its coverage. The benefit of the neural network model is that it can tag any word because the PLM of the context and gloss encoder is capable of embedding any word in context. We integrate the neural network model into the rule-based model, so that when the rule-based model fails to make a prediction due to a word being not defined in the lexicon(s), the neural model can step in as a back-off model.

\section{Experimental Setup}
For the neural network model, we test 4 different variants with different PLMs; 2 English PLMs from \citet{weller2025seq}, and 2 Multilingual PLMs from \citet{marone2025mmbert}, which are outlined in Table \ref{table:neural_network_model_details}. These PLMs were chosen as they have been shown to perform best in their size category, and are completely open including the data they were trained on, and the multilingual models have been pre-trained on all the languages we are evaluating on. Only the Multilingual PLMs will be tested on the non-English datasets.

Each of the neural networks are fine-tuned on the English silver labelled training dataset from section \ref{section:silver_training_data}, using the loss function detailed in section \ref{section:neural_models}. The models are trained with early stopping on the validation split of the silver labelled training dataset, whereby we track the accuracy of the model predicting the correct sense rather than the 3 negative examples. We checkpoint the models each time it has trained on 20\% of the training samples. All models were trained with a batch size of 64 and a learning rate of $1\mathrm{e}{-5}$. Due to computing resource constraints we only managed to train the \textbf{N\textsubscript{EngS}} model completely, as in the early stopping constraint stopped the model training. For all other models, we used the best performing checkpoint. The number of epochs the models trained for can be seen in table \ref{table:model_training_epochs_validation_accuracy}.

\begin{table}[!ht]
    \centering
    \begin{tabular}{c|c|c}
    \hline
        Model & Epochs & Validation accuracy (\%) \\
    \hline
        \textbf{N\textsubscript{EngS}} & 3 & 99.39 \\
        \textbf{N\textsubscript{EngB}} & 3 & 99.67 \\
        \textbf{N\textsubscript{MulS}} & 2.2 & 99.62 \\
        \textbf{N\textsubscript{MulB}} & 1.36 & 99.63 \\
    \hline
    \end{tabular}
    \caption{The neural network model, the number of epochs it trained for on the silver labelled training data, and the validation accuracy on the validation split of the silver labelled training data.}
    \label{table:model_training_epochs_validation_accuracy}
\end{table}

As outlined in section \ref{section:rule-based-models}, we only use one rule-based model per language that will be represented as \textbf{R} in the results. The hybrid models using the rule-based model per language will be represented as \textbf{H\textsubscript{neural model}}, e.g. the hybrid model using the small multilingual neural network will be \textbf{H\textsubscript{MulS}}.

All models are evaluated on top-n accuracy, whereby top-1 only considers the first tag generated by the model, and top-5 considers the first 5 tags generated by the model. In this evaluation, a single tag can include Multi Tag Membership for both the true tag and the predicted. In these cases, for it to be correct, the true and predicted tags must be identical. For example, if the predicted tag is \textit{F2/O2} and the correct tag is \textit{F2/O1}, then the prediction is considered wrong. This way of evaluating means that, for the neural network based predictions, they will always fail to predict the Multi Tag Membership true tags, because it has not been trained to predict these type of tags.

When evaluating on the non-English datasets, this will be a cross-lingual evaluation setup, as the neural network based models have only been fine-tuned on English data for this WSD task.

\begin{table}[!ht]
    \centering
    \begin{tabular}{c|c}
    \hline
        Name & Parameters (M) \\
    \hline
        \textbf{N\textsubscript{EngS}} & 16.8 \\
        \textbf{N\textsubscript{EngB}} & 68.1 \\
        \textbf{N\textsubscript{MulS}} & 140.5 \\
        \textbf{N\textsubscript{MulB}} & 306.9 \\
    \hline
    \end{tabular}
    \caption{The name of the neural network models where English Small (EngS) and English Base (EngB) use the base model \textit{Ettin-Enc-17m} and \textit{Ettin-Enc-68m} respectively from \citet{weller2025seq} and Multilingual Small (MulS) and Mulitlingual Base (MulB) use the base model \textit{MMBERT small} and \textit{MMBERT base} respectively from \citet{marone2025mmbert}. The number of parameters in millions (M).}
    \label{table:neural_network_model_details}
\end{table}

\section{Results}

The top-n accuracy results are shown in Table \ref{table:top_n_accuracy_results}\footnote{The Irish language results are not directly comparable with the rule based semantic tagging results of \citet{czerniak-ui-dhonnchadha-2024-towards}, this is due to; 1. different pre-processing, in this paper, of the evaluation data whereby punctuation is not treated as a labelled token and affixed symbols such as \%,@,m,f,+, and - were removed from the USAS tags, and 2. different evaluation metrics have been used, the "Correctness" for all tokens metric from \citet{czerniak-ui-dhonnchadha-2024-towards} is comparable to top-1 accuracy within this paper.}, where we can see across all languages and values of \textit{n} that the the best model is either neural network model or the hybrid model, demonstrating the benefit of training a neural network model.

There is a large difference in the evaluation data results in table \ref{table:top_n_accuracy_results} and the validation accuracy results in table \ref{table:model_training_epochs_validation_accuracy} on the silver labelled data. This is due to the fact the validation accuracy is performed on an easier task in that it is predicting the most likely sense between 1 positive sense and 3 negative senses, whereas the evaluation task is identifying the correct sense from all of the USAS tag senses, therefore very high validation accuracy is expected.

For $n=1$, all but the Chinese rule-based models performed better than the neural network models, but in all cases the hybrid models performed best. The same findings can be seen for $n=5$ except that the English rule-based model performs worse than the neural network models.

For Chinese, it can be seen that, for both $n=1$ and $n=5$, the neural network models (\textbf{N\textsubscript{MulS}} and \textbf{N\textsubscript{MulB}}) outperformed both the rule-based and hybrid models due to the poor performance of the rule-based model. This is also the case for English when $n=5$. This highlights a drawback of the hybrid model whereby if the recall of the rule-based model is high and the precision is low then the hybrid model will be limited by the rule-based model's performance. It will be less likely to fall back to the neural network model due to the rule-based model's high recall. This, we hypothesise is the reason why there is little difference in performance between the rule-based model and the hybrid model for English for both values of $n$.

In general, we find that the larger the neural network model the better the performance. However, for English the two base models, \textbf{N\textsubscript{EngB}} and \textbf{N\textsubscript{MulB}}, their performances are comparable, showing that a smaller language specific model can be as performant as a larger multilingual model.

\begin{table}[!ht]
    \centering
    \begin{tabular}{c|c}
    \hline
    Language     & Pre-Training Tokens (B) \\
    \hline
    Welsh     & 2.2 \\
    Irish     & 3 \\
    Finnish     & 25.5 \\
    Chinese     & 139.2 \\
    English     & 803.3 \\
    \hline
    \end{tabular}
    \caption{Number of pre-training tokens in billions (B), in which both multilingual PLMs were pre-trained. Statistics is taken from table 9 of \citet{marone2025mmbert}.}
    \label{table:multilingual-pre-training-token-statistics}
\end{table}

\begin{table*}[!ht]
    \centering
    \begin{tabular}{cccccc|ccccc}
 & \multicolumn{5}{c|}{n=1} & \multicolumn{5}{c}{n=5} \\
 & Chinese & English & Finnish & Irish & Welsh & Chinese & English & Finnish & Irish & Welsh \\
 \hline
\textbf{R} & 32.6 & \underline{72.4} & 58.4 & \underline{56.6} & 70.6 & 43.6 & 81.8 & 64.0 & \underline{62.1} & 73.2 \\
\hline
\textbf{N\textsubscript{EngS}} & - & 66.4 & - & - & - & - & 87.6 & - & - & - \\
\textbf{N\textsubscript{EngB}} & - & 70.1 & - & - & - & - & \underline{90.0} & - & - & - \\
\hdashline
\textbf{H\textsubscript{EngS}} & - & \textbf{72.5} & - & - & - & - & 81.9 & - & - & - \\
\textbf{H\textsubscript{EngB}} & - & \textbf{72.5} & - & - & - & - & 82.0 & - & - & - \\
\hline
\textbf{N\textsubscript{MulS}} & \underline{42.2} & 66.0 & 15.8 & 28.5 & 21.7 & \underline{66.3} & 88.9 & 32.8 & 47.6 & 40.8 \\
\textbf{N\textsubscript{MulB}} & \textbf{47.9} & 70.2 & 25.9 & 35.6 & 42.0 & \textbf{70.4} & \textbf{90.1} & 42.4 & 51.6 & 56.4 \\ 
\hdashline
\textbf{H\textsubscript{MulS}} & 39.8 & \textbf{72.5} & \underline{59.1} & \textbf{57.1} & \underline{71.3} & 55.6 & 82.0 & \underline{65.8} & \textbf{63.3} & \underline{75.5} \\
\textbf{H\textsubscript{MulB}} & 39.8 & \textbf{72.5} & \textbf{60.3} & \textbf{57.1} & \textbf{72.4} & 56.3 & 82.0 & \textbf{67.3} & \textbf{63.3} & \textbf{75.9} \\
\hline
\end{tabular}

    \caption{Top \textit{n} accuracy results for all models and languages. The best performing results per language per value of \textit{n} is denoted in \textbf{bold} and the second best is \underline{underlined}. The results are divided by a horizontal line by rule-based method (\textbf{R}), then English only neural and hybrid models, and then multilingual neural and hybrid models. The neural and hybrid models are divided within the English and multilingual by the dashed horizontal line. The dash (-) denotes a result that is not applicable, which in all cases is using a English neural model for non-English texts.}
    \label{table:top_n_accuracy_results}
\end{table*}

We can see that the performance of the multilingual neural network models is significantly higher for Chinese and English. We assume this is, in part, due to the huge amount of Chinese and English data available for pre-training the model, as shown in Table \ref{table:multilingual-pre-training-token-statistics}. The implication of this result is that it is worth investigating: a) using a more language specific PLM to enhance the performance of the tagger, b) fine-tuning the PLM to the target specific languages.

\section{Conclusion and Future Work}

In this work, we have created the first neural network English and multilingual semantic taggers that have been trained specifically for the USAS tagset without any manually annotated data using only English silver labelled data created from a English rule-based tagger. We have also demonstrated how a hybrid neural/rule-based method can be created.

Through the most extensive evaluation so far, with respect to number of languages, we have shown that either the hybrid or neural network based models across all languages outperform the existing rule-based methods, showing the advantages of these new neural network or hybrid based models for USAS tagging. In future work, we will evaluate the hybrid model further to disentangle the performance gains and losses from the neural network model and rule-based model respectively. We will also evaluate the neural network and rule-based models on tokens that are defined in the rule-based model's lexicon(s) so that we can better isolate the disambiguate performance of the models, rather than evaluating both coverage and disambiguate, in addition this evaluation will inform us of whether neural network models can go beyond the performance of the rule-based model that it learnt from.

Our results show that even though we have not fine-tuned on any language other than English the neural network models have performed well for Chinese, which we believe is due to the amount of Chinese in the pre-training data that PLM was pre-trained on (see table \ref{table:multilingual-pre-training-token-statistics}), which can also explain the reason why they performed worse for the low resourced languages (Finnish, Irish, and Welsh). Future work could explore the affect of using a language specific PLM across more languages, as we found in this work that a smaller English specific PLM (\textbf{N\textsubscript{EngB}}) performed similarly to a larger multilingual PLM (\textbf{N\textsubscript{MulB}}).

In this work, we have released the first open manually annotated corpus for USAS semantic tagging in Chinese. Our initial qualitative comparison of the rule-based Chinese model with the manually annotated Chinese corpus revealed that Chinese measure words, dates, and time expressions are semantic categories where the model can be improved. These categories will be monitored and systematically evaluated in future work.

We have also released the first silver labelled dataset in English that has been shown to be useful for training both mono and multilingual neural networks for USAS tagging. Future work could explore extending the silver labelled datasets into other languages using the same methodology of using existing rule-based USAS tagger, which will allow exploration of fine-tuning of language specific neural networks models.

All of the resources created or used in this paper have been made available as open resources. The neural network and hybrid models are also available to use within the PyMUSAS framework\footnote{\url{https://ucrel.github.io/pymusas/}}, which is an easy to use Python semantic tagging framework that already contains the rule based models. 

\section{Limitations}

This work acknowledges that creating the silver labelled training data that we used to train the neural network based models does require an existing rule-based model. The effect of the accuracy of such rule-based model on the performance of the trained neural network model is currently unknown and could be explored in the future as this is likely to be a limitation for future work which may create silver labelled datasets for low resource languages with potentially less performant rule-based models. However, given our successful cross lingual experiments based on English, this may be unnecessary.

With respect to future work of low resource languages, such as Irish and Welsh, we understand that the amount of data to create a silver labelled corpus would be a limitation as compared to a higher resourced language where the amount of text available for these languages is much less.

\section{Acknowledgements}

This research was partially funded by the 4D Picture Project\footnote{\url{https://4dpicture.eu/}}. Andrew Moore was also funded by the University Centre for Computer Corpus Research on Language (UCREL) at Lancaster University, UK\footnote{\url{https://ucrel.lancs.ac.uk/}}. The 4D Picture project research leading to these results has received funding from the EU research and innovation programme HORIZON Europe 2021 under grant agreement 101057332 and by the Innovate UK Horizon Europe Guarantee Programme, UKRI Reference Number 10041120.

We are grateful to the UCREL research group at Lancaster University for providing us access to the Hex computer cluster \cite{UcrelHex} where the experiments were processed.

\section{Bibliographical References}\label{sec:reference}

\bibliographystyle{lrec2026-natbib}
\bibliography{references}

\section{Language Resource References}
\label{lr:ref}
\bibliographystylelanguageresource{lrec2026-natbib}
\bibliographylanguageresource{language-references.bib}

\section*{Appendix A. Additional Data Details}
\label{appendix:additional_data_details}

\begin{figure}
    \centering
    \includesvg[scale=0.37]{figures/training_data_label_distribution_heatmap.svg}
    \caption{Probability of a USAS label within the silver labelled training data, training split, for each distribution.}
    \label{figure:probability_distribution_of_usas_in_silver_data}
\end{figure}

\end{document}